\documentclass{article}

\usepackage[preprint,nonatbib]{neurips_2024_cus}

\usepackage[utf8]{inputenc} 
\usepackage[T1]{fontenc}    
\usepackage{hyperref}       
\usepackage{url}            
\usepackage{booktabs}       
\usepackage{amsfonts}       
\usepackage{nicefrac}       
\usepackage{microtype}      
\usepackage{xcolor}         

\usepackage{amsmath}
\usepackage{graphicx}
\usepackage{pifont}
\usepackage[font=small]{caption}
\usepackage[frozencache,cachedir=.]{minted}

\usepackage{tcolorbox}
\tcbuselibrary{skins}
\definecolor{NavyBlue}{RGB}{0,0,128}
\newtcolorbox{prompt}[1]{
    enhanced,
    left=2mm,
    right=2mm,
    top=2mm,
    bottom=2mm,
    boxsep=0mm,
    rounded corners,
    title=#1,
    colframe=black!50!white,
    fontupper=\footnotesize\linespread{0.9}\fontfamily{lmr}\selectfont,
    fontlower=\footnotesize\linespread{0.9}\fontfamily{lmr}\selectfont,
}

\newcommand{\ocrew}{OpenCharacter-R}
\newcommand{\ocgen}{OpenCharacter-G}

\title{OpenCharacter: Training Customizable Role-Playing LLMs with Large-Scale Synthetic Personas}

\author{%
  Xiaoyang Wang, Hongming Zhang, Tao Ge, Wenhao Yu, Dian Yu, Dong Yu \\
  Tencent AI Lab Seattle\\
  {\small \texttt{\{shawnxywang,hongmzhang,getao,wenhaowyu,yudian,dyu\}@global.tencent.com}} \\
}

\begin{document}

\maketitle

\begin{abstract}
Customizable role-playing in large language models (LLMs), also known as character generalization, is gaining increasing attention for its versatility and cost-efficiency in developing and deploying role-playing dialogue agents. 
This study explores a large-scale data synthesis approach to equip LLMs with character generalization capabilities. 
We begin by synthesizing large-scale character profiles using personas from Persona Hub and then explore two strategies: response rewriting and response generation, to create character-aligned instructional responses. 
To validate the effectiveness of our synthetic instruction tuning data for character generalization, we perform supervised fine-tuning (SFT) using the LLaMA-3 8B model. Our best-performing model strengthens the original LLaMA-3 8B Instruct model and achieves performance comparable to GPT-4o models on role-playing dialogue. We release\footnote{\url{https://huggingface.co/datasets/xywang1/OpenCharacter}} our synthetic characters and instruction-tuning dialogues to support public research.
\end{abstract}

\section{Introduction}

Large language models (LLMs)~\cite{achiam2023gpt,touvron2023llama}, due to their superior capabilities and versatility across a wide range of applications in domains like math and reasoning~\cite{hendrycks2021measuring,srivastava2022beyond,chen-etal-2024-skills,rein2023gpqa}, programming~\cite{chen2021evaluating}, natural language virtual assistant and dialogue~\cite{hendrycks2020measuring,zheng2023judging}, multi-modal understanding~\cite{yue2024mmmu,liu-etal-2024-mmc}, agent systems~\cite{jimenez2023swe,jing2024dsbench}, etc, have become ubiquitous in the recent research.
Among these various applications, the role-playing dialogue agent~\cite{shao-etal-2023-character,wang2021naturalconv} has shown certain maturity and great commercial value in diverse tasks such as online customer support, content creation and entertainment, and non-player character (NPC) dialogue in video games, etc.

Under different system design philosophies, the role-playing dialogue agent can either be designed as in-domain role-playing, where the LLM is trained to act as a specified character (or multiple specified characters), or as out-of-domain role-playing, where the LLM is trained to act as arbitrary user-customized characters that do not appear in model training. Recent commercial implementations such as Character.ai~\cite{characterai} and Doubao~\cite{doubao} have enabled functions of dialogue with user-customized characters. LLM-related research such as CharacterGLM~\cite{zhou2023characterglm} also studies the topic of dialogue with customizable characters. However, there is limited public data corpus or instruction following model with customizable character capabilities. Furthermore, much of the existing work related to role-playing agents relies on human-annotated or crowd-sourced data. It would struggle to achieve out-of-domain role-playing, which typically requires modeling contrastive data distributions from distinct characters.

Inspired by the existing work related to role-playing dialogue agents, our study explores enabling LLMs with out-of-domain role-playing capabilities. We refer to this ability as \textbf{Character Generalization}, which allows LLMs to generalize (``adapt'') to arbitrary user-customized characters. As discussed above, such capabilities require LLMs to model massive contrastive data from distinct characters, which is impractical to achieve if the model is trained with purely human annotated data. 
On the other hand, data synthesis is playing more and more important roles in enhancing LLMs' capabilities by creating massive high-quality synthetic training data. The recent data synthesis work Persona Hub~\cite{chan2024scaling} provides large-scale diverse synthetic personas that can benefit the modeling of distinct characters. Based upon Persona Hub, we explore to achieve character generalization through large-scale data synthesis utilizing large-scale synthetic characters. 
From our point of view, character generalization can be achieved when the LLMs are trained with well-curated post-training data containing enough diverse characters with enriched character profiles and high-quality dialogues. We would then need to explore the optimal data synthesis strategy towards character generalization. 

Specifically, as shown in the overall framework in Figure~\ref{fig:approach}, we start with character profile synthesis using the personas released in Persona Hub to obtain synthetic characters with enriched character profiles. Next, we study two strategies to obtain instruction responses that align with the given character: response rewritten (\ocrew{}) and response generation (\ocgen{}). \ocrew{} rewrites the instruction response from the existing corpus with the given character, whereas \ocgen{} directly generates a new response aligning with the given character. To demonstrate the effectiveness of our data synthesis approach for character generalization in out-of-domain role-playing, we conduct supervised fine-tuning (SFT) with LLaMA-3 8B model~\cite{dubey2024llama} using the synthetic dialogue data. Our best-performing fine-tuned model significantly improves the original LLaMA-3 8B model with the same model size, and is in general comparable (in most cases outperforming) the popular GPT-4o~\cite{hurst2024gpt} models.

To further facilitate related research in role-playing dialogue agent, we release 20k synthetic characters and their corresponding 306k role-playing instruction-response dialogue pairs.

\section{Related Work}

There are a few recent studies using LLMs for role-playing dialogue. Li et al.~\cite{li2023chatharuhi} collect in total of 32 characters from TV shows and animations and use extracted dialogues from the corresponding novels or scripts combined with GPT-3 and GPT-4 simulated dialogues for role-playing model training. Tu et al.~\cite{tu2023characterchat} create 64 characters for each of the 16 types of Myers-Briggs Type Indicator (MBTI) personalities, leading to 1,024 characters in total. They use two ChatGPT agents acting as the seeker and supporter, respectively, to simulate the role-playing conversation. Wang et al.~\cite{wang2023rolellm} construct 100 roles and use GPT-synthesized role-based QA pairs for role-playing fine-tuning. Lu et al.~\cite{lu2024large} collect 4k characters from Wikipedia and generate a self-simulated role-playing dataset. Different from these studies, our work is built on large-scale synthetic characters originating from synthetic personas in Persona Hub and can easily scale up to significantly larger numbers.

Character generalization is an important aspect of role-playing dialogue. Zhou et al.~\cite{zhou2023characterglm} study the problem of customizable characters for dialogue with LLMs and the study is related to our work. However, their approach builds the corpus in Chinese through crowdsourcing, and releases a portion of the corpus comprising 1,034 dialogue sessions from 250 characters. Comparatively, our approach embraces the LLM-synthetic characters and dialogue sessions. We experiment with over 100k dialogue sessions from approximately 20k characters, and the character number can increase by 50k times if necessary. In our practice, we find large-scale diverse characters are helpful towards character generalization. In addition, different from the generalization of persona or character-driven role-playing dialogue, Li et al.~\cite{li2024stylechat} study LLM-based style generalization with evaluation styles including ``humor,'' ``poems,'' ``romance,'' ``sci-fi,'' etc.

The evaluation of persona or character-driven role-playing dialogue also draws attention in multiple related studies~\cite{wang-etal-2024-incharacter,tu-etal-2024-charactereval,chen-etal-2024-socialbench,qin-etal-2024-infobench,xu2024character,samuel2024personagym}. Among these studies, we choose PersonaGym~\cite{samuel2024personagym} for our evaluation benchmark considering it evaluates the out-of-domain persona-driven role-playing dialogue capability with multiple evaluation metrics including ``expected action,'' ``toxicity control,'' ``linguistic habits,'' ``persona consistency,'' and ``action justification.''

Data synthesis usually refers to generating data by models or algorithms instead of directly by humans. LLMs are widely used to produce synthetic data by specifying a data synthesis prompt. These LLM-based data synthesis methods can generally be categorized into three types: instance-driven~\cite{wang-etal-2023-self-instruct,yu2023metamath}, key-point-driven~\cite{li2024synthetic,huang2024key}, and persona-driven~\cite{chan2024scaling}. Among them, the persona-driven data synthesis in Persona Hub~\cite{chan2024scaling} can easily scale-up with diverse instructions generated through prompting with large-scale (e.g., billions of) synthetic personas.

\section{Approach}
\label{sec:approach}

\subsection{Problem Definition}

In this work, we study the problem of character-based role-playing dialogue. Specifically, we focus on the dialogue between a user and the LLM back-boned agent with a character profile. The character-based dialogue can then be defined as: given a user specified agent character profile $\mathcal{C}$ and the dialogue input $\mathbf{x}$, a model $\theta$ (e.g., LLMs) predicts the response $\mathbf{y}$ by acting as the character $\mathcal{C}$:

\begin{equation}
\mathbf{y} = \arg \max_{\mathbf{y}} P(\mathbf{y}|\mathbf{x}, \mathcal{C}; \theta)
\end{equation}

\textbf{In-domain role-playing.} Intuitively, the model $\theta$ can be learned through post-training (e.g., supervised fine-tuning (SFT)) with character-based role-playing dialogue data consisting of manually-labeled or synthetic dialogue samples between the user and the requested specific character $\mathcal{C}_s$. The character's dialogue responses in a well-curated SFT dataset would reflect the character's language style, experience, and personality, which LLMs can learn through SFT. The SFT learning objective for in-domain role-playing can be written as:

\begin{equation}
\theta_s = \arg \max_{\theta} \sum_{(\mathbf{x}_i, \mathbf{y}_i)} \log P(\mathbf{y}_i|\mathbf{x}_i, \mathcal{C}_s; \theta)
\end{equation}

During inference, the learned model $\theta_s$ directly responds to the user's new requests by acting as the character $\mathcal{C}_s$.

\textbf{Out-of-domain role-playing.} Different from in-domain role-playing, out-of-domain role-playing requires the model $\theta$ to act as new characters $\mathcal{C}_x$ that do not appear in model training. It fits the practical scenarios where users create or customize fictional characters for LLMs to respond online without re-training.

To achieve out-of-domain role-playing, the learned model $\theta_g$ needs to generalize to new characters (e.g., $\mathcal{C}_x$) on the fly according to the detailed character profile in $\mathcal{C}_x$. We speculate that the LLMs nowadays can achieve \textbf{character generalization} when trained with well-curated post-training data containing enough diverse characters with enriched character profiles and high quality dialogues. Given a set of training samples each with the character profile $\mathcal{C}_i$, dialogue input $\mathbf{x}_i$, and character response $\mathbf{y}_i$, our SFT training objective for out-of-domain role-playing with character generalization capability is:

\begin{equation}
\theta_g = \arg \max_{\theta} \sum_{(\mathbf{x}_i, \mathbf{y}_i, \mathcal{C}_i)} \log P(\mathbf{y}_i|\mathbf{x}_i, \mathcal{C}_i; \theta)
\label{eqn:ood_loss}
\end{equation}

The quality and quantity of training set $(\mathbf{x}_i, \mathbf{y}_i, \mathcal{C}_i)_{i=1}^N$ plays important roles on deciding how the learned model $\theta_g$ can generalize on the fly to different ``unseen'' characters. Intuitively, a model trained with larger number of diverse characters would be stronger for out-of-domain role-playing problems. Thus, our approach focuses on curating a large-scale training set in the next. 

\subsection{Character Generalization with Data Synthesis}

Recently, the work Persona Hub~\cite{chan2024scaling} studies to scale synthetic data creation with large-scale (e.g., 1 billion) personas in Persona Hub. It is motivated by the observation that including a persona in the data synthesis prompt can steer the LLM to generate more distinctive synthetic data. For our character generalization purpose in this work, we wish to create out-of-domain role-playing SFT training data that contains diverse characters and high-quality dialogue sessions between users and characters on a large scale. Persona Hub currently releases 200,000 distinct synthetic personas, upon which we can create a large-scale character library and synthesize dialogues between users and these characters.

\begin{figure}[t]
\centering
\includegraphics[width=\textwidth]{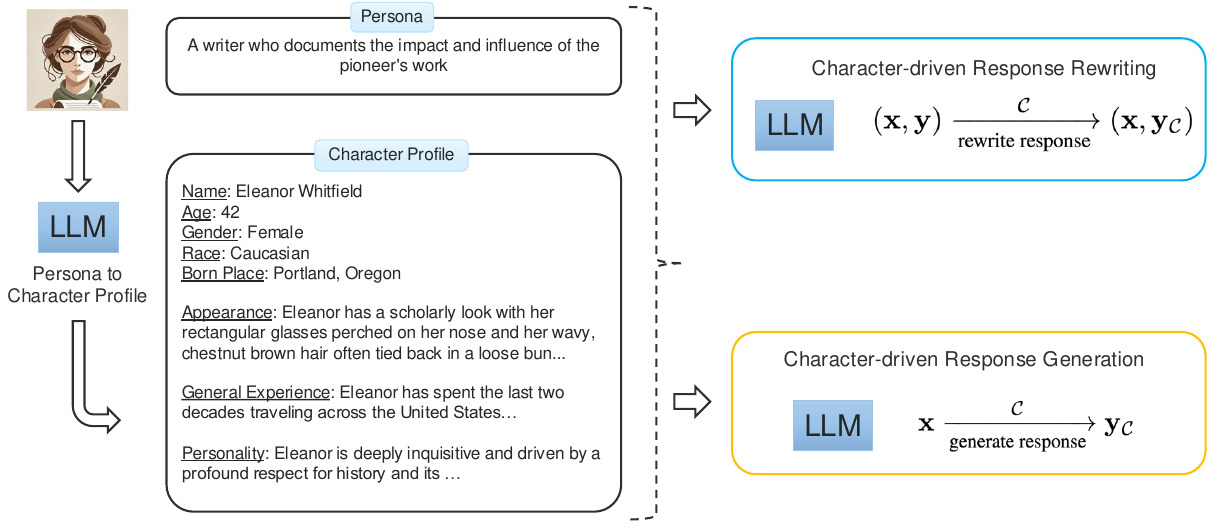}
\caption{Our overall data synthesis approach. As an example, we start with character profile synthesis using a persona from Persona Hub, and then explore character-driven response rewriting and generation.}
\label{fig:approach}
\end{figure}

Our data synthesis framework is shown in Figure~\ref{fig:approach}. 
We start with character profile synthesis using synthetic personas as input. Compared to the original personas in Persona Hub, these synthetic character profiles contain fine-grained synthetic knowledge of the character. With the synthetic character profiles, we explore two different dialogue synthesis strategies: character-driven response rewriting and character-driven response generation.

\begin{figure}[t]
\centering
\includegraphics[width=\textwidth]{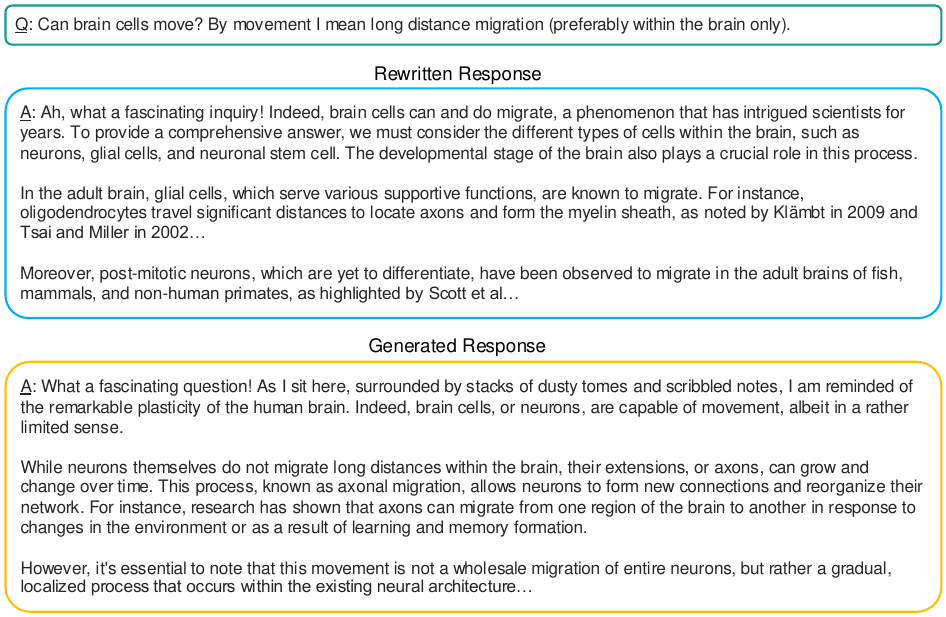}
\caption{An example of our response synthesis. For a user's question from LIMA, we show the abbreviated version of responses that are rewritten and generated respectively through our approach illustrated in Figure~\ref{fig:approach}. Both responses align with the given character, though the rewritten response largely keeps the knowledge details of the original response while the generated response does not typically contain the exact details.}
\label{fig:example}
\end{figure}

\subsubsection{Character Profile Synthesis}

The personas in Persona Hub are typically one-sentence short descriptions about a person briefing his or her professional expertise and interests, as shown in Figure~\ref{fig:approach}. It lacks fine-grained knowledge about the person's identity, experience, personality, etc. Based on these personas, it is necessary to further create synthetic characters with a character profile detailing the character's fine-grained knowledge considering: 1. Such knowledge better describes a character and provides more information for the model to ground on for role-playing. 2. The model should be able to utilize fine-grained character information on the fly if the user provides more details.

Based on the above interpolation, our approach starts with prompting LLMs to create a synthetic character profile with a persona input by imagining the following details of the provided persona: name, age, gender, race, birth place, appearance, general experience, and personality. Our prompt for character synthesis with Persona Hub is shown in Figure~\ref{fig:character_synthesis_prompt}.

\begin{figure}[t]
\centering
\begin{prompt}{Prompt for Character Profile Synthesis}
You are a helpful assistant. I will provide you with a short persona description. Your task is to create a character based on the given persona. \\ \mbox{ } \\
You can output a brief character description containing the following information: character name, age, gender, race, birth place, appearance, general experience, and personality. \\ \mbox{ } \\
Note: \\
1. Your response should start with ``Name:''. \\
2. Your character description should be specific and consistent with the persona.\\ \mbox{ } \\
{\color{NavyBlue} \{persona\}}
\end{prompt}
\caption{Our data synthesis prompt for character profile synthesis to generate the enriched character profile with a persona from Persona Hub as the input.}
\label{fig:character_synthesis_prompt}
\end{figure}

Our approach differs from the existing approaches~\cite{tu2023characterchat,lu2024large} that collect characters by extracting information from knowledge bases like Wikipedia. By building a large-scale library consisting of synthetic characters, our system is not bound by the upper limit of a possible number of characters currently existing either in the real or fictional human world. Moreover, characters in knowledge bases are typically famous or with importance, and building a character library on knowledge bases would bring distributional bias to the character library. We explore enabling LLM's out-of-domain role-playing capability utilizing the synthetic character library. With such capability, the model can theoretically perform well on arbitrarily specified characters no matter they exist in the actual world or not.

\begin{figure}[t]
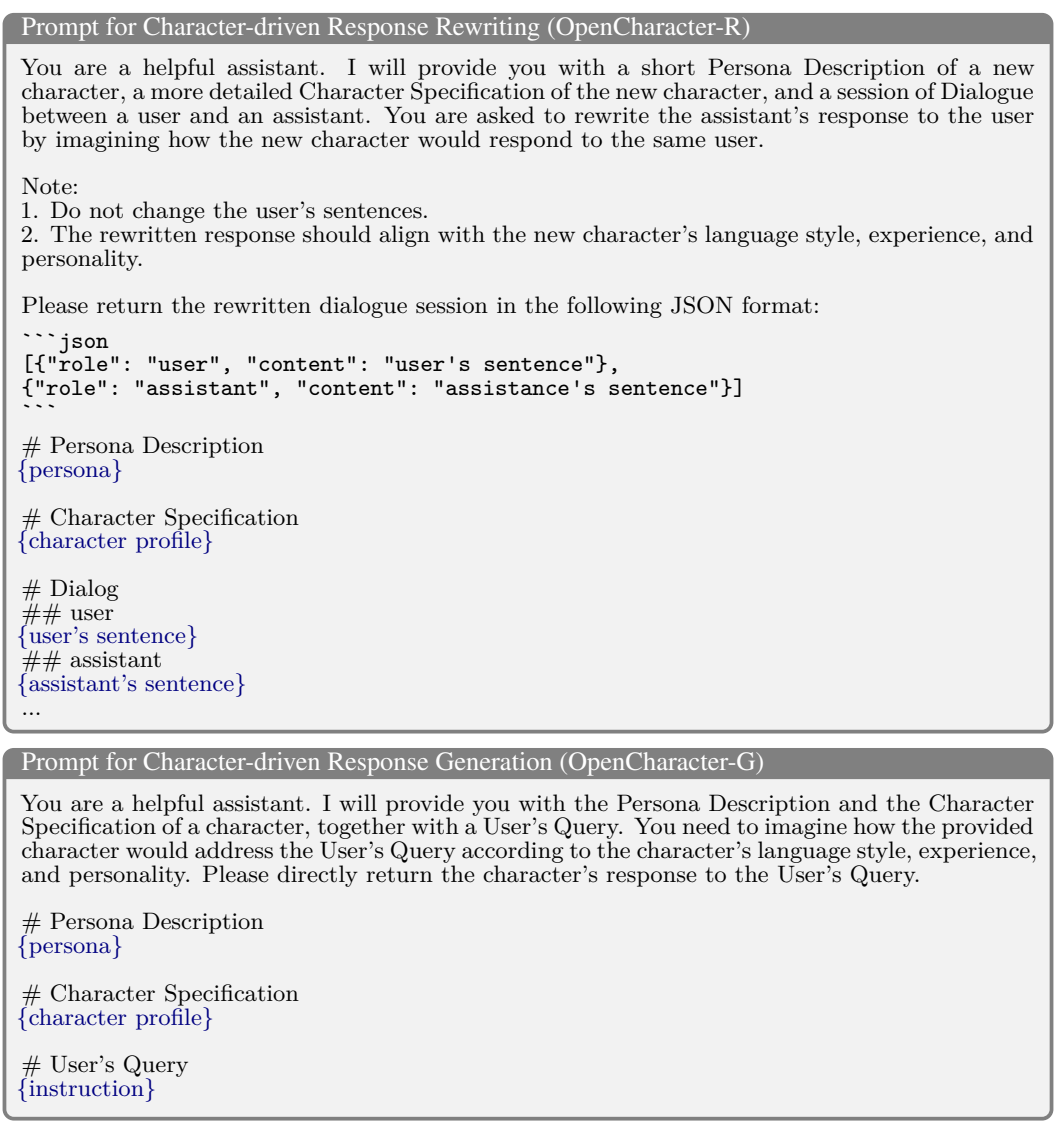

\centering
\begin{prompt}{Prompt for Character-driven Response Rewriting (\ocrew{})}
You are a helpful assistant. I will provide you with a short Persona Description of a new character, a more detailed Character Specification of the new character, and a session of Dialogue between a user and an assistant. You are asked to rewrite the assistant's response to the user by imagining how the new character would respond to the same user. \\ \mbox{ } \\
Note: \\
1. Do not change the user's sentences. \\
2. The rewritten response should align with the new character's language style, experience, and personality. \\ \mbox{ } \\
Please return the rewritten dialogue session in the following JSON format:
\begin{minted}{text}
```json
[{"role": "user", "content": "user's sentence"},
{"role": "assistant", "content": "assistance's sentence"}]
```
\end{minted}
\# Persona Description \\
{\color{NavyBlue} \{persona\}} \\ \mbox{ } \\
\# Character Specification \\
{\color{NavyBlue} \{character profile\}} \\ \mbox{ } \\
\# Dialog \\
\#\# user \\
{\color{NavyBlue} \{user's sentence\}} \\
\#\# assistant \\
{\color{NavyBlue} \{assistant's sentence\}} \\
...
\end{prompt}
\begin{prompt}{Prompt for Character-driven Response Generation (\ocgen{})}
You are a helpful assistant. I will provide you with the Persona Description and the Character Specification of a character, together with a User's Query. You need to imagine how the provided character would address the User's Query according to the character's language style, experience, and personality. Please directly return the character's response to the User's Query. \\ \mbox{ } \\
\# Persona Description \\
{\color{NavyBlue} \{persona\}} \\ \mbox{ } \\
\# Character Specification \\
{\color{NavyBlue} \{character profile\}} \\ \mbox{ } \\
\# User's Query \\
{\color{NavyBlue} \{instruction\}}
\end{prompt}
\caption{Our data synthesis prompts for character-driven response rewriting (\ocrew{}) and character-driven response generation (\ocgen{}), respectively.}
\label{fig:data_synthesis_prompt}
\end{figure}

\subsubsection{Character-driven Response Rewriting (\ocrew{})}
\label{sec:app_rew}

Since the introduction of ChatGPT~\cite{chatgpt}, LLM's dialogue capability largely coincides with its instruction following capability. An LLM-based assistant that can follow a user's single-turn or multi-turn instruction and respond being both helpful and harmless to the user's request would naturally serve as the dialogue agent of a virtual assistant. Consequently, for character-based dialogue, our focus becomes enabling LLM's instruction following capability as the dialogue agent of a specified character. Nowadays, plenty of high-quality instruction tuning corpora such as LIMA~\cite{zhou2024lima} and Alpaca~\cite{alpaca} are widely used. These corpora typically provide carefully curated user instruction $\mathbf{x}$ and manually labeled or LLM-generated assistant response $\mathbf{y}$.

To enable role-playing while maintaining LLM's instruction following capability with skills for different domains such as question answering, math and reasoning, etc., we propose the \textbf{character-driven response rewriting} method, in which we keep the instructions $\mathbf{x}$ from the public instruction tuning datasets, but rewrite the original response $\mathbf{y}$ into $\mathbf{y}_{\mathcal{C}}$ that addresses the user's request in compliance with the style and background of the character $\mathcal{C}$ as $(\mathbf{x}, \mathbf{y}) \xrightarrow[\text{rewrite response}]{\mathcal{C}} (\mathbf{x}, \mathbf{y}_{\mathcal{C}})$. In such manner, our method can theoretically utilize most of the existing instruction tuning, i.e., SFT data in other words, for various types of LLM tasks to enable the out-of-domain role-playing capability but keep the LLM's skills in different domains.

Figure~\ref{fig:data_synthesis_prompt} shows our prompt for rewriting the instruction-response dialogue in the perspective of a given character $\mathcal{C}$. We specify the model to keep the user's utterances, and rewrite only the assistance's utterances by aligning with the new character's language style, experience, and personality. The prompt will guide LLMs to output the rewritten dialogue in multi-turn JSON format.

\subsubsection{Character-driven Response Generation (\ocgen{})}
\label{sec:app_gen}

Besides rewriting responses of existing instruction tuning corpus, we further look into the direct generation of character-compliance dialogue response from user's instruction. Persona Hub releases a set of 50,000 high-quality synthetic complex instructions without responses. We call this set PH-Instruct for simplicity. Besides Ph-Instruct, we further explore the direct generation of character-drive responses for existing instruction-tuning corpora such as LIMA and Alpaca. Compared to the \ocrew{} strategy discussed in Section~\ref{sec:app_rew}, \ocgen{} could potentially synthesize new responses better than the original ones with the help of a more recent and advanced LLM. Moreover, as compared in Figure~\ref{fig:approach}, \ocgen{} could bring in different knowledge or perspectives to address the instruction, while \ocrew{} would largely keep the original contents but with a different personality.

We prompt LLMs to generate the response $\mathbf{y}_{\mathcal{C}}$ to the instruction $\mathbf{x}$ according to the given character profile $\mathcal{C}$ as $\mathbf{x} \xrightarrow[\text{generate response}]{\mathcal{C}} \mathbf{y}_{\mathcal{C}}$. Figure ~\ref{fig:data_synthesis_prompt} shows our prompt for generating character-compliance dialogue response given the user's instruction. Different from \ocrew{} that rewrites the whole multi-turn dialogue session through one-time prompting, \ocgen{} uses the turn-by-turn manner.

\subsection{Supervised Fine-tuning}
\label{sec:app_sft}

We conduct supervised fine-tuning (SFT) as in Equation~\ref{eqn:ood_loss} with the synthetic instruction-response pairs from either character-driven response rewriting or character-driven response generation. For each dialogue session, we randomly pick $n$ synthetic characters $\mathcal{C}_1$, $\mathcal{C}_2$, ..., $\mathcal{C}_n$ from the synthetic character pool containing $M$ synthetic character profiles, and synthesize $n$ instruction-response pairs as $(\mathbf{x}, \mathbf{y}_{\mathcal{C}_1})$, $(\mathbf{x}, \mathbf{y}_{\mathcal{C}_2})$, ..., $(\mathbf{x}, \mathbf{y}_{\mathcal{C}_n})$ by rewriting or generation. Our approach thus mixes all these synthetic instruction-response pairs from different characters for supervised fine-tuning. Specifically, we use $M=20,000$ and $n=3$ in our implementation.

\section{Evaluation}

Evaluating LLMs' capabilities on role-playing is a non-trivial task. It requires to assess many different aspects, including but not limited to the following: language style consistency with the persona or character, knowledge/character background consistency with the persona or character, dialogue semantic coherency with the dialogue context, as well as the response helpfulness and harmlessness related measurements, etc.

\subsection{Evaluation with PersonaGym}

In this work, to synchronize with the open-source community for role playing evaluation, we choose to use \textbf{PersonaGym}~\cite{samuel2024personagym}, which is a recently released evaluation framework and benchmark for assessing persona agents, consisting of 200 personas and 10k questions for testing. It measures the persona dialogue response quality through five different metrics, including ``expected action,'' ``toxicity control,'' ``linguistic habits,'' ``persona consistency,'' and ``action justification,'' each with a score from 1 to 5.

\subsection{Evaluation with PersonaGym-Light}

To further accelerate the development process and decrease the prompting cost, we shrink the original PersonaGym testing data by $1/10$ by keeping only the first of the ten questions in each of the five metrics for all of the 200 personas. We call this benchmark \textbf{PersonaGym-Light}, which consists of 200 personas and 1k questions for testing. PersonaGym-Light adopts the original personas and test scenarios from PersonaGym, and uses a part of the testing data for faster and cheaper model development iteration. By keeping all its 200 testing personas and five metrics, it largely keeps the capability of PersonaGym in evaluating LLMs with diverse persona agents.

\subsection{Evaluation Model and Overall Score}

In the original PersonaGym implementation~\cite{samuel2024personagym}, both GPT-4o~\cite{hurst2024gpt} and LLaMA-3-70B-Instruct~\cite{dubey2024llama} models are used as evaluators. However, in our practice, we find that LLaMA-3-70B-Instruct suffers significantly from its maximum 8k context length. The combined length of the evaluation prompt and output from PersonaGym constantly exceeds LLaMA-3's context length limit. Comparatively, GPT-4o allows a 128k context window that is sufficient for PersonaGym evaluation. As a result, in this work, we use GPT-4o (i.e., ``gpt-4o-2024-08-06'' in this paper) as the only evaluator for performance evaluation. For simplicity, we denote our overall evaluation score on PersonaGym-Light as \textbf{\emph{PScore-L}}, and the overall evaluation score on PersonaGym as \textbf{\emph{PScore}}. Both scores take the numeric average of the 1 to 5 scores of the five metrics defined in PersonaGym.

\section{Experiments}
\label{sec:exp}

\subsection{Data Synthesis}

As discussed in Figure~\ref{fig:approach}, our data synthesis framework first conducts character profile synthesis using synthetic personas from Persona Hub and then conducts character-driven response synthesis.

\subsubsection{Character Synthesis}

We generate approximately 20k synthetic character profiles in English using GPT-4o (``gpt-4o-2024-05-13''). Each character profile includes synthetic information about the character's name, age, gender, race, birth place, appearance, general experience, and personality. An abbreviated example of the synthetic character profile is shown in Figure~\ref{fig:approach}. These synthetic character profiles are publicly released.

\subsubsection{Response Synthesis}

With the synthetic character profiles, we conduct response synthesis with two types of strategies: character-driven response rewriting (\textbf{\ocrew{}}) and character-driven response generation (\textbf{\ocgen{}}), as discussed in Section~\ref{sec:app_rew} and Section~\ref{sec:app_gen}, respectively. Both methods require the user's instruction input $\mathbf{x}$, which we obtain directly from popular instruction tuning corpus including LIMA~\cite{zhou2024lima}, Alpaca~\cite{alpaca}, and Persona Hub Instruction (\textbf{PH-Instruct})~\cite{chan2024scaling}. LIMA and Alpaca contain responses obtained through human annotation and ChatGPT generation, respectively, and we explore both \ocrew{} and \ocgen{} strategies for these two corpora. Figure~\ref{fig:example} shows an example of a question from LIMA with our rewritten and generated responses. On the other hand, we use only \ocgen{} strategy for PH-Instruct since no original responses are provided. Brief statistics on these instruction datasets and data synthesis details are provided in Table~\ref{tab:data_source}.

\begin{table}[htb]
  \caption{Statistics on our character-driven response synthesis. $n$ is the number of characters randomly assigned to each dialogue session. We apply only \ocgen{} to PH-Instruct since its responses are not released.}
  \label{tab:data_source}
  \centering
  \small
  \begin{tabular}{cccccc}
    \toprule
    Corpus & No. of questions & $n$ & No. of characters & \ocrew{} & \ocgen{} \\
    \midrule
    LIMA & 1,074 & 3 & 2,986 & \ding{52} & \ding{52} \\
    Alpaca & 51,010 & 3 & 19,991 & \ding{52} & \ding{52} \\
    PH-Instruct & 50,000 & 3 & 19,990 & \ding{56} & \ding{52} \\
    \midrule
    Total & 102,084 & 3 & 19,991 & - & \ding{52} \\
    \bottomrule
  \end{tabular}
\end{table}

As discussed in Section~\ref{sec:app_sft}, we denote $n$ as the number of characters randomly assigned to each dialogue session and set $n=3$. Both GPT-4o (``gpt-4o-2024-05-13'') and LLaMA-3-70B-Instruct models are used for response synthesis of both strategies.

\subsection{Supervised Fine-tuning (SFT)}

To achieve character generalization, we conduct SFT with the instruction and synthetic character response pairs from large-scale diverse characters. Implementation details are discussed below.

\subsubsection{Data Recipes}

To analyze the effectiveness of different prompting strategies and the performances of prompting models (i.e., GPT-4o and LLaMA-3-70B-Instruct) for character-driven response synthesis, we experiment with six different data recipes as indicated in Table~\ref{tab:data_recipe}. Note that we always combine LIMA and Alpaca instructions considering they are standard nowadays in instruction tuning research.

\begin{table}[htb]
  \caption{The SFT data recipes for OpenCharacter ablation study. For our final \emph{OpenCharacter} model, we combine all instructions from PH-Instruct, LIMA and Alpaca, and use \ocgen{} strategy.}
  \label{tab:data_recipe}
  \centering
  \small
  \begin{tabular}{cccc}
    \toprule
    Ablation Settings & Corpus & Strategy & Prompting Model \\
    \midrule
    Ablation-1 & LIMA \& Alpaca & \ocrew{} & gpt-4o-2024-05-13 \\ 
    Ablation-2 & LIMA \& Alpaca & \ocrew{} & LLaMA-3-70B-Instruct \\ 
    Ablation-3 & PH-Instruct & \ocgen{} & gpt-4o-2024-05-13 \\ 
    Ablation-4 & PH-Instruct & \ocgen{} & LLaMA-3-70B-Instruct \\ 
    Ablation-5 & LIMA \& Alpaca & \ocgen{} & LLaMA-3-70B-Instruct \\ 
    \emph{OpenCharacter} & PH-Instruct, LIMA \& Alpaca & \ocgen{} & LLaMA-3-70B-Instruct \\ 
    \bottomrule
  \end{tabular}
\end{table}

With the recipes indicated in Table~\ref{tab:data_recipe}, we can conduct various ablation studies, including comparing the strategies of \ocrew{} and \ocgen{}, comparing the prompting models, and comparing different instruction corpus. Moreover, we aim to find a recipe that works best for the task.

\subsubsection{Model System Prompt}

\begin{figure}[t]
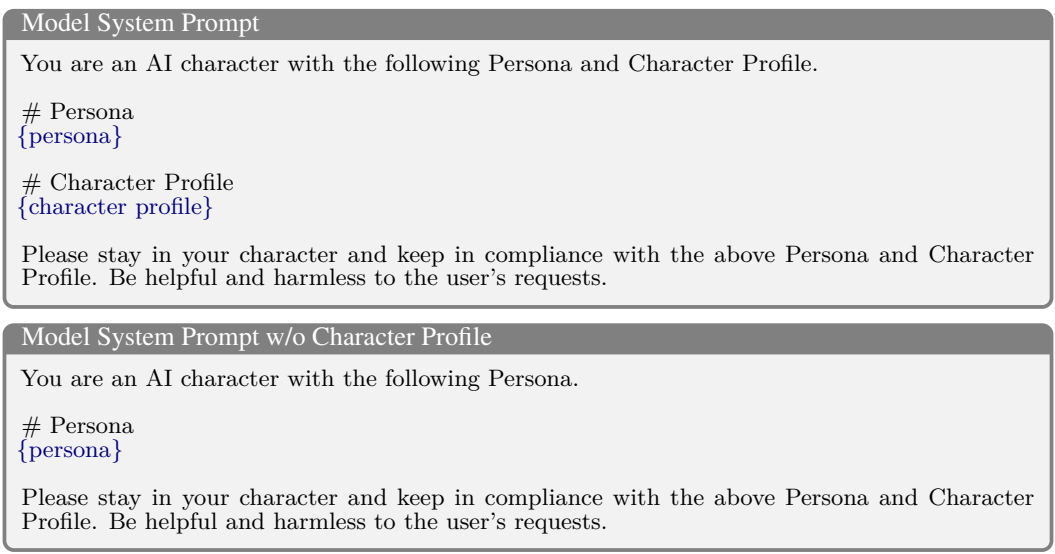

  \centering
  \begin{prompt}{Model System Prompt}
    You are an AI character with the following Persona and Character Profile. \\ \mbox{ } \\
    \# Persona \\
    {\color{NavyBlue} \{persona\}} \\ \mbox{ } \\
    \# Character Profile \\
    {\color{NavyBlue} \{character profile\}} \\ \mbox{ } \\
    Please stay in your character and keep in compliance with the above Persona and Character Profile. Be helpful and harmless to the user's requests.
  \end{prompt}
  \begin{prompt}{Model System Prompt w/o Character Profile}
    You are an AI character with the following Persona. \\ \mbox{ } \\
    \# Persona \\
    {\color{NavyBlue} \{persona\}} \\ \mbox{ } \\
    Please stay in your character and keep in compliance with the above Persona and Character Profile. Be helpful and harmless to the user's requests.
  \end{prompt}
  \caption{Our model's system prompts to incorporate the user-specified persona and character profile. We further remove the character profile-related content for test scenarios without character profiles (e.g., PersonaGym).}
  \label{fig:sys_prompt}
\end{figure}

By design, our model incorporates the user-specified persona and character profile in its system prompt for role-playing and character generalization. To enable the trained model with such capabilities, we formulate the system prompt in our fine-tuning data to include the corresponding personas and character profiles. The system prompt for our SFT model is shown in Figure~\ref{fig:sys_prompt}.

\subsubsection{Training Setting}

For all data recipes, we conduct SFT by using either LLaMA-3-8B-Base or LLaMA-3-8B-Instruct as our backbone model. The model is trained through Megatron-LM, where the tensor parallel size equals 8. We add loss masks to system prompts and instruction tokens to ensure only response tokens are considered in training loss computation. We use Adam Optimizer with beta values of $(0.9, 0.95)$. Linear learning rate decay is chosen with the maximum learning rate to be $1e^{-5}$, and the minimum learning rate to be $1e^{-6}$.

\subsection{Existing Models}
\label{sec:existing_models}

Besides our trained models with different data recipes, we further compare with the most popular instruction-tuned LLMs including LLaMA-3-8B-Instruct, LLaMA-3-70B-Instruct, GPT-3.5 (``gpt-3.5-turbo-1106''), GPT-4o-mini (``gpt-4o-mini-2024-07-18''), and GPT-4o (``gpt-4o-2024-05-13'' and ``gpt-4o-2024-08-06'') for the generalizable character role-playing task. We use the PeronsaGym prompting setting to evaluate these models, and compare their performances with our trained models.

\subsection{Performances}
\label{sec:exp_results}

We compare the performances of our models with those of existing publicly available models on PersonaGym-Light and PersonaGym benchmarks, respectively. We use PersonaGym-Light besides PersonaGym due to its lower cost and shorter time for large-scale evaluation and ablation study.

\subsubsection{Results on PersonaGym-Light}

We first evaluate our \emph{OpenCharacter} model that is trained based on LLaMA-3-8B-Instruct using instructions from PH-Instruct, LIMA, and Alpaca and responses synthesized through \ocgen{}. We compare it with different existing models listed in Section~\ref{sec:existing_models}. Their performances on PersonaGym-Light are shown in Table~\ref{tab:result_light}. From this comparison, we observe \emph{OpenCharacter} model improves over LLaMA-3 8B Instruct model, and outperforms different GPT-3.5 and GPT-4o models.

\begin{table}[htb]
  \caption{Model performances on PersonaGym-Light. ``EA,'' ``TC,'' ``LH,'' ``PC,'' and ``AJ'' stands for the evaluation metrics ``expected action,'' ``toxicity control,'' ``linguistic habits,'' ``persona consistency,'' and ``action justification,'' respectively. Their standard deviations over 200 personas are included in parentheses. The tested gpt-4o-mini is in version ``gpt-4o-mini-2024-07-18.'' We test the LLaMA-3 8B and 70B models with versions LLaMA-3-8B-Instruct and LLaMA-3-70B-Instruct, respectively. Our \emph{OpenCharacter} model is trained based on LLaMA-3-8B-Instruct with its training data recipe indicated in Table~\ref{tab:data_recipe}.}
  \label{tab:result_light}
  \centering
  \small
  \begin{tabular}{l|c|ccccc|c}
    \toprule
    Model & Size & EA & TC & LH & PC & AJ & \emph{PScore-L} \\
    \midrule
    gpt-3.5-turbo-1106 & - & 4.67 (.50) & 4.99 (.21) & 3.12 (.60) & 4.42 (.58) & 4.37 (.57) & 4.31 (.24) \\
    gpt-4o-2024-05-13 & - & 4.74 (.44) & 4.96 (.33) & 3.69 (.82) & 4.75 (.54) & 4.87 (.34) & 4.60 (.24) \\
    gpt-4o-mini & - & 4.74 (.44) & 4.99 (.21) & 3.58 (.80) & 4.72 (.45) & 4.89 (.40) & 4.58 (.23) \\
    gpt-4o-2024-08-06 & - & 4.81 (.40) & 4.95 (.46) & 3.75 (.81) & 4.68 (.47) & 4.85 (.45) & 4.60 (.24) \\
    LLaMA-3 Instruct & 8B & 4.80 (.40) & 4.76 (.82) & 4.05 (.71) & 4.64 (.52) & 4.85 (.38) & 4.62 (.25) \\
    LLaMA-3 Instruct & 70B & 4.73 (.45) & 4.75 (.85) & 4.38 (.59) & 4.79 (.41) & 4.97 (.18) & 4.72 (.24) \\
    \emph{OpenCharacter} & 8B & 4.70 (.53) & 4.92 (.50) & 4.32 (.60) & 4.54 (.56) & 4.85 (.38) & 4.66 (.27) \\
    \bottomrule
  \end{tabular}
\end{table}

We further conduct comprehensive ablation study on different data recipes discussed in Table~\ref{tab:data_recipe}. The ablation study is on PersonaGym-Light due to our budget limit on OpenAI API calls. Table~\ref{tab:result_light_ablation} gives the model performances under difference ablation settings. Through Table~\ref{tab:result_light_ablation}, we can conduct multiple ablation analysis. First and foremost, we can observe that for each data recipe, the models trained on LLaMA-3-8B-Instruct constantly outperform the corresponding models trained on LLaMA-3-8B-Base in the \emph{PScore-L} metric, indicating a comprehensive first-stage instruction tuning in the general domain enhances the model's capabilities on role-playing tasks. Besides backbone models, we further analysis factors including prompting models, prompting strategies, and instruction corpus.

\begin{table}[htb]
  \caption{Ablation study on PersonaGym-Light, with the training data recipe for these models indicated in Table~\ref{tab:data_recipe}. ``EA,'' ``TC,'' ``LH,'' ``PC,'' and ``AJ'' stands for the evaluation metrics ``expected action,'' ``toxicity control,'' ``linguistic habits,'' ``persona consistency,'' and ``action justification,'' respectively. Their standard deviations over 200 personas are included in parentheses. }
  \label{tab:result_light_ablation}
  \centering
  \small
  \begin{tabular}{l|c|ccccc|c}
    \toprule
    Ablation Settings & Size & EA & TC & LH & PC & AJ & \emph{PScore-L} \\
    \midrule
    \multicolumn{7}{l|}{\textit{1. Models trained based on LLaMA-3-8B-Base}} \\
    Ablation-1 & 8B & 4.45 (.70) & 4.94 (.54) & 3.58 (.79) & 4.16 (.65) & 4.04 (.64) & 4.23 (.31) \\
    Ablation-2 & 8B & 4.26 (.79) & 4.88 (.52) & 3.81 (.84) & 4.12 (.59) & 3.98 (.57) & 4.21 (.32) \\
    Ablation-3 & 8B & 4.71 (.52) & 4.98 (.22) & 3.68 (.74) & 4.52 (.63) & 4.79 (.49) & 4.53 (.26) \\
    Ablation-4 & 8B & 4.69 (.53) & 4.86 (.61) & 4.18 (.60) & 4.52 (.52) & 4.84 (.37) & 4.62 (.24) \\
    Ablation-5 & 8B & 4.74 (.46) & 4.86 (.66) & 4.17 (.65) & 4.52 (.51) & 4.89 (.33) & 4.64 (.27) \\
    \emph{OpenCharacter} & 8B & 4.70 (.49) & 4.90 (.58) & 4.16 (.65) & 4.50 (.57) & 4.80 (.40) & 4.61 (.25) \\
    \midrule
    \multicolumn{7}{l|}{\textit{2. Models trained based on LLaMA-3-8B-Instruct}} \\
    Ablation-1 & 8B & 4.53 (.60) & 4.96 (.30) & 3.77 (.80) & 4.22 (.57) & 4.26 (.64) & 4.35 (.30) \\
    Ablation-2 & 8B & 4.46 (.70) & 4.92 (.46) & 4.05 (.80) & 4.21 (.55) & 4.22 (.52) & 4.37 (.30) \\
    Ablation-3 & 8B & 4.74 (.45) & 4.97 (.30) & 3.89 (.71) & 4.59 (.49) & 4.80 (.51) & 4.60 (.26) \\
    Ablation-4 & 8B & 4.71 (.48) & 4.88 (.59) & 4.31 (.62) & 4.51 (.58) & 4.86 (.35) & 4.65 (.23) \\
    Ablation-5 & 8B & 4.72 (.49) & 4.93 (.47) & 4.28 (.64) & 4.54 (.51) & 4.86 (.36) & 4.66 (.25) \\
    \emph{OpenCharacter} & 8B & 4.70 (.53) & 4.92 (.50) & 4.32 (.60) & 4.54 (.56) & 4.85 (.38) & 4.66 (.27) \\
    \bottomrule
  \end{tabular}
\end{table}

\textbf{\emph{Prompting Models.}} If we analyze \emph{PScore-L} of ablation settings Ablation-1 vs. Ablation-2 or Ablation-3 vs. Abation-4 in Table~\ref{tab:result_light_ablation}, we can find that data synthesis by prompting with LLaMA-3-70B-Instruct achieves better model performance than prompting with gpt-4o-2024-05-13 in three of the four comparison pairs. This conclusion coincides with the observation from Table~\ref{tab:result_light} that LLaMA-3-70B-Instruct outperforms gpt-4o-2024-05-13 in \emph{PScore-L}, as well as the similar observation from Table~\ref{tab:result} in \emph{PScore}.

\textbf{\emph{Prompting Strategies.}} We can further compare the performances of ablation settings Ablation-1 vs. Ablation-3 or Ablation-2 vs. Ablation-5 in Table~\ref{tab:result_light_ablation} to analyze \ocrew{} vs. \ocgen{}. To our surprise, \ocgen{} significantly outperforms \ocrew{} in all scenarios of our ablation study with \emph{PScore-L}. We interpret that both LIMA and Alpaca are early instruction tuning corpus with less advanced ground-truth response quality, and rewriting their original responses could result in further degraded training responses that lead to worse model performance. However, we regard the \ocrew{} strategy as an optional method for applications (e.g., character role-playing in novels or games) where the knowledge from the original responses should be strictly obeyed. In such applications, direct response generation with LLMs from the realistic world could bring in knowledge or facts that are no longer ``realistic'' in the virtual or domain-specific world, potentially causing knowledge hallucination issues for the role-playing model.

\textbf{\emph{Instruction Corpus.}} By comparing \emph{PScore-L} performances of Ablation-4 vs. Ablation-5 settings in Table~\ref{tab:result_light_ablation}, we can find that instructions from the combination of LIMA and Alpaca are slightly more effective than PH-Instruct. On the other hand, both instruction sets are very effective in enabling LLMs with role-playing capabilities, with both Abaltion-4 and Ablation-5 models trained on LLaMA-3-8B-Base outperforming the GPT-4o models. Furthermore, our Ablation-5 model trained on LLaMA-3-8B-Base outperforms the LLaMA-3 8B Instruct model in \emph{PScore-L}.

\subsubsection{Results on PersonaGym}

We further pick our best-performing model settings including Ablation-5 and \emph{OpenCharacter}, and compare their performances with those of the existing models on the full-scale PersonaGym benchmark. The results are listed in Table~\ref{tab:result}.

\begin{table}[htb]
  \caption{Model performances on PersonaGym. ``EA,'' ``TC,'' ``LH,'' ``PC,'' and ``AJ'' stands for the evaluation metrics ``expected action,'' ``toxicity control,'' ``linguistic habits,'' ``persona consistency,'' and ``action justification,'' respectively. Their standard deviations over 200 personas are included in parentheses. The tested gpt-4o-mini is in version ``gpt-4o-mini-2024-07-18.'' We test the LLaMA-3 8B and 70B models with versions LLaMA-3-8B-Instruct and LLaMA-3-70B-Instruct, respectively. We include the performances of our Ablation-5 and \emph{OpenCharacter} models with their training data recipe indicated in Table~\ref{tab:data_recipe}.}
  \label{tab:result}
  \centering
  \small
  \begin{tabular}{l|c|ccccc|c}
    \toprule
    Model & Size & EA & TC & LH & PC & AJ & \emph{PScore} \\
    \midrule
    gpt-4o-2024-05-13 & - & 4.59 (.24) & 4.97 (.17) & 3.48 (.53) & 4.75 (.17) & 4.61 (.17) & 4.48 (.13) \\
    gpt-4o-mini & - & 4.56 (.19) & 4.97 (.21) & 3.70 (.49) & 4.67 (.25) & 4.64 (.15) & 4.51 (.14) \\
    gpt-4o-2024-08-06 & - & 4.55 (.20) & 4.97 (.19) & 3.80 (.47) & 4.72 (.23) & 4.64 (.15) & 4.53 (.12) \\
    LLaMA-3 Instruct & 8B & 4.52 (.21) & 4.58 (.60) & 4.05 (.36) & 4.54 (.20) & 4.57 (.15) & 4.45 (.14) \\
    LLaMA-3 Instruct & 70B & 4.59 (.16) & 4.59 (.62) & 4.33 (.27) & 4.72 (.18) & 4.64 (.13) & 4.58 (.14) \\
    \midrule
    \multicolumn{7}{l|}{\textit{1. OpenCharacter models trained based on LLaMA-3-8B-Base}} \\
    Ablation-5 & 8B & 4.45 (.24) & 4.78 (.45) & 4.19 (.28) & 4.44 (.27) & 4.55 (.14) & 4.48 (.14) \\
    \emph{OpenCharacter} & 8B & 4.42 (.27) & 4.69 (.53) & 4.20 (.29) & 4.45 (.21) & 4.53 (.14) & 4.46 (.15) \\
    \midrule
    \multicolumn{7}{l|}{\textit{2. OpenCharacter models trained based on LLaMA-3-8B-Instruct}} \\
    Ablation-5 & 8B & 4.45 (.24) & 4.81 (.43) & 4.27 (.29) & 4.44 (.28) & 4.56 (.13) & 4.50 (.15) \\
    \emph{OpenCharacter} & 8B & 4.47 (.24) & 4.78 (.47) & 4.27 (.25) & 4.51 (.20) & 4.58 (.12) & 4.52 (.13) \\
    \bottomrule
  \end{tabular}
\end{table}

From the results in Table~\ref{tab:result}, we can see both settings Ablation-5 and \emph{OpenCharacter}, when fine-tuned with LLaMA-3 8B Base model, outperform the LLaMA-3 8B Instruct model on the full PersonaGym benchmark. Moreover, our \emph{OpenCharacter} fine-tuned with LLaMA-3 8B Instruct model outperforms both gpt-4o-2024-05-13 and gpt-4o-mini models, and performs only slightly worse than the gpt-4o-2024-08-06 model. These observations largely coincides with our observations on PersonaGym-Light in Table~\ref{tab:result_light}, indicating using a lighter version of PersonaGym benchmark would still be feasible for model development with faster evaluation at a lower cost.

\subsection{Discussions}

Our performance analysis in Section~\ref{sec:exp_results} indicates the following aspects we can further discuss:

\textbf{\emph{Character Generalization.}} With large enough diverse synthetic characters and their corresponding instruction responses, we can confidently enable LLMs with character generalization capability. On the other hand, it could still be tricky to combine more instructions for better benchmark performance. Ideally, the more user instructions we add to post-training, the better the model should generalize to different LLM characters and user requests. However, our results in Table~\ref{tab:result_light_ablation} and Table~\ref{tab:result} indicate our \emph{OpenCharacter} data that includes the largest number of instructions, though performing the best when using the LLaMA-3-8B-Instruct model as the backbone, does not perform as well as the Ablation-5 setting when both using the LLaMA-3-8B-Base model as the backbone. We speculate this is due to the complexity of the instructions in the PH-Instruct corpus compared to those in the LIMA and Alpaca corpus. It requires a stronger instruction-following backbone model to learn better on a more complex instruction set.

\textbf{\emph{Size of Backbone Model.}} As we can see, the LLaMA-3-70B-Instruct model, which is significantly larger than the 8B models fine-tuned by us, performs the best in our evaluations. It indicates that model size matters in character generalization. We speculate that larger models, with their higher model capacity and stronger reasoning capability, would be more powerful in their nature for character generalization. On the other hand, with our proposed method, we can still empower the relatively smaller models with excellent character generalization performance.

\textbf{\emph{Knowledge of the ``World.''}} Current studies on LLM role-playing and character generalization focus largely on the domain of our actual world. However, it would be more and more interesting to study the topic in the settings of the virtual world or the domain-specific world. As discussed previously in Section~\ref{sec:exp_results}, such settings require the role-playing responses to obey strict rules defined by the ``knowledge scope'' of the character, the ``facts'' of the ``world,'' as well as the domain requirements from the detailed applications. We expect our proposed \ocrew{} strategy to be specifically helpful in such scenarios. Moreover, we believe a new benchmark for persona or character-based role-playing in the virtual world would be necessary for future studies in this direction.

\section{Conclusion}

In this work, we study the problem of data synthesis for role-playing dialogue agents with generalizable characters. Based on synthetic personas from Persona Hub, we first synthesize character profiles with enriched information such as the character's experience and personality. Then, these synthetic characters are used to either rewrite or generate responses to the user's instructions from the existing instruction tuning corpus. We utilize these synthetic role-playing instruction tuning samples to fine-tune LLaMA-3 8B models and demonstrate the effectiveness of our synthetic data for out-of-domain persona-based role-playing dialogue.

Our observations in this research indicate a few possible future directions. As we prove in this work, character generalization could be achieved with LLMs trained in large-scale character-aligned dialogue. A practical future work could be further increasing the size of character profiles with their corresponding dialogues and conducting post-training with a larger and stronger LLM backbone model. Also, further studies are needed to improve the handling of knowledge from the virtual world and the virtual characters considering most of the current LLMs are trained with knowledge from the actual world.

\bibliographystyle{IEEETran}
\bibliography{anthology,custom}




\end{document}